\documentclass[10pt,twocolumn,letterpaper]{article}
\usepackage{cvpr}
\usepackage{times}
\usepackage{epsfig}
\usepackage{graphicx}
\usepackage{amsmath}
\usepackage{amssymb}
\usepackage{threeparttable}
\usepackage{verbatim}
\usepackage{caption}
\usepackage{subfigure}
\newcommand{\grad}{\ensuremath{^{\circ}}}
\usepackage{authblk}
\usepackage{epstopdf}

\usepackage[breaklinks=true,bookmarks=false]{hyperref}

\cvprfinalcopy 

\def\cvprPaperID{****} 

\begin{document}

\title{  Boundary-Aware Network for Fast and High-Accuracy Portrait Segmentation  }

\author{Xi Chen}
\author{Donglian Qi}
\author[]{Jianxin Shen}
\affil{College of Electrical Engineering, Zhejiang University, Hangzhou 310027, China}
\affil{ \tt\small \{ xichen\_zju, qidl, j\_x\_shen \}@zju.edu.cn}

\maketitle

\begin{abstract}
   Compared with other semantic segmentation tasks, portrait segmentation requires both higher precision 
   and faster inference speed. However, this problem has not been well studied in previous works. In this paper, we propose a light-weight network architecture, called Boundary-Aware Network (BANet) which selectively extracts detail 
   information in boundary area to make high-quality segmentation output with real-time( $\ge$ 25FPS) speed. In addition, we design a new
   loss function called refine loss which supervises the network with image level gradient information.
   Our model is able to produce finer segmentation results which has richer details than annotations.

\end{abstract}

\section{Introduction}

Portrait segmentation has been widely applied in many practical scenes such as mobile phone photography, video surveillance and
image preprocess of face identification. Compared with other semantic segmentation tasks, portrait segmentation has higher requirement for
both accuracy and speed. However, this task has not been well studied in previous works. PFCN+ \cite{shen2016automatic} uses prior information
to enhance segmentation performance, BSN \cite{du2017boundary} designs a boundary kernel to emphasize the importance of boundary area. These methods
get high mean IoU on PFCN+ \cite{shen2016automatic} dataset, but they are extremely time consuming because they use large backbones. What's more,
their segmentation results lack detail information such as threads of hairs and clothe boundary details. Although automatic image matting models such as \cite{shen2016deep} 
\cite{zhu2017fast} are able to obtain fine details, but preparing high-quality alpha matte is much more complicated and time-consuming than annotate segmentation
targets. In this paper, we propose an novel method that can generate high-accuracy portrait segmentation result with real-time speed.  

\par In recent years, many deep learning based semantic segmentation models like\cite{chen2018deeplab}\cite{zhao2017pyramid}\cite{zhang2018exfuse}\cite{yu2018learning} have reached state-of-the-art performance. However, when it comes to portrait segmentation, none of these works produces fine boundary. The problem of losing fine boundary details is mainly caused by two reasons. 
\par
\begin{figure}[htbp]
\centering

\subfigure{
\begin{minipage}[t]{0.33\linewidth}
\centering
\includegraphics[width=1in]{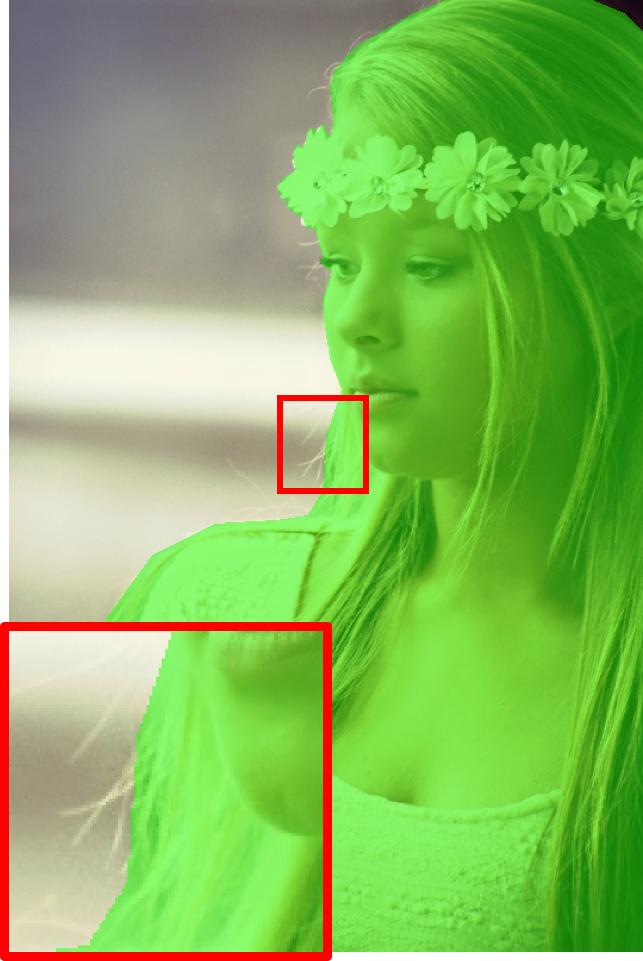}
\end{minipage}%
}%
\subfigure{
\begin{minipage}[t]{0.33\linewidth}
\centering
\includegraphics[width=1in]{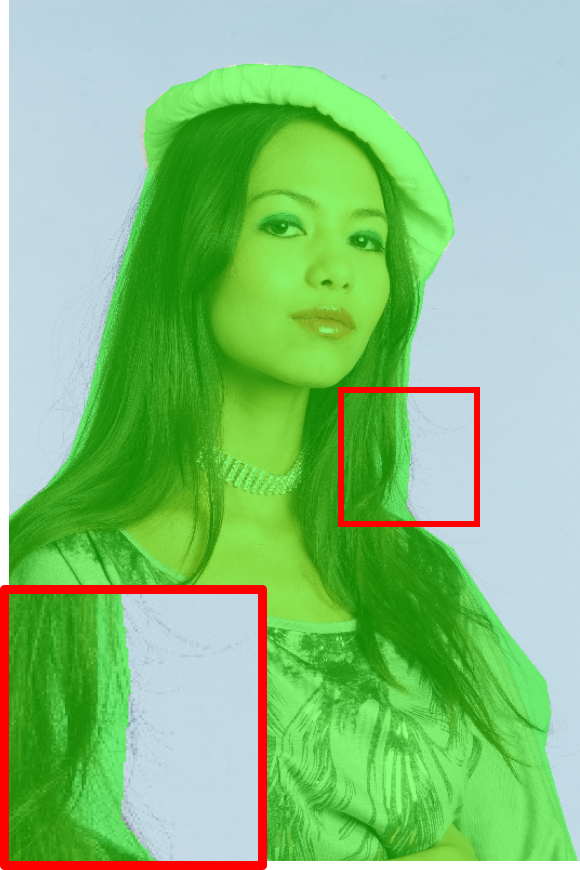}
\end{minipage}
}%
\subfigure{
\begin{minipage}[t]{0.33\linewidth}
\centering
\includegraphics[width=1in]{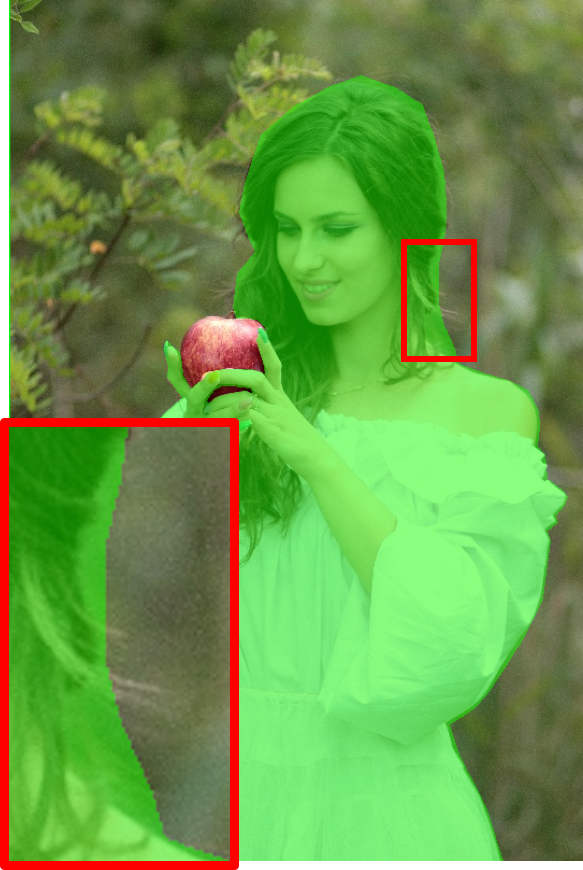}
\end{minipage}
}%
\centering
\caption{ Polygon Annotations of Supervise.ly  }
\label{fig_sup}
\end{figure}

 Firstly, performance of deep learning based method is quite 
dependent on the quality of training data. While segmentation targets are usually  annotated manually by polygons or produced by KNN-matting \cite{chen2013knn}. Therefore, it is almost impossible to annotate fine details such as hairs. Supervise.ly dataset is currently the largest and finest human segmentation dataset, some examples are presented as Fig. \ref{fig_sup}. It shows that although Supervise.ly is the finest dataset, annotations are still coarse on boundary areas. Coarse annotation causes inaccurate supervision, which makes it hard for neural network to learn a accurate feature representation.        
\par Secondly, Although previous portrait segmentation methods\cite{shen2016automatic}\cite{du2017boundary} have proposed some useful pre-process methods or loss functions, they still use traditional semantic segmentation models like FCN\cite{long2015fully}, Deeplab\cite{chen2018deeplab} and PSPnet\cite{zhao2017pyramid} as their baseline. The problem is that the output size of those models is  usually 1/4 or 1/8 of input image size, thus it is impossible to preserve as much detail information as input image. U-Net\cite{ronneberger2015u} fuses more low-level information but it is oriented for small sized medical images, it is hard to train on high-resolution portrait images on account of its huge parameter number. 
Although traditional semantic segmentation models achieves state-of-the-art performance on Pascal VOC\cite{everingham2010pascal}, COCO Stuff\cite{caesar2016coco} and CitySpace\cite{cordts2016cityscapes} datasets, they are not suitable for portrait segmentation. Traditional segmentation task aims at handling issues of intra-class consistency and the inter-class distinction among various object classes in complicated scenes. While portrait segmentation is a two-class task which requires fine details and fast speed.  Hence,
portrait segmentation should be considered as an independent task and some novel strategies should be proposed.

\begin{figure}[htbp]
\centering
\subfigure[Image]{
\begin{minipage}[t]{0.33\linewidth}
\centering
\includegraphics[width=1in]{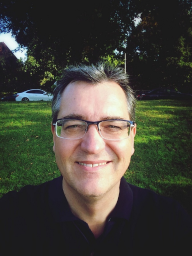}
\includegraphics[width=1in]{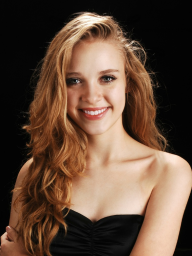}
\includegraphics[width=1in]{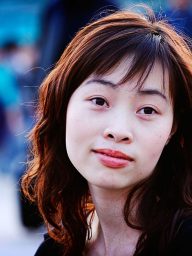}

\end{minipage}%
}%
\subfigure[Our Result]{
\begin{minipage}[t]{0.33\linewidth}
\centering
\includegraphics[width=1in]{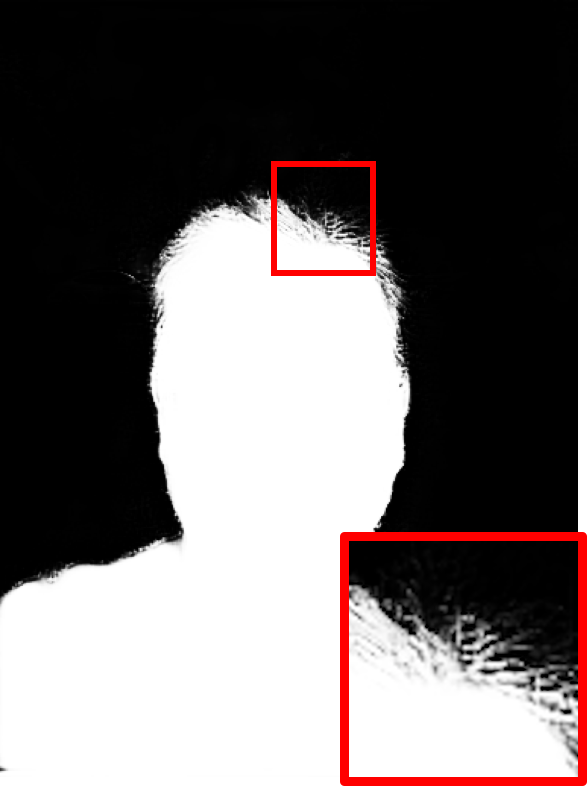}
\includegraphics[width=1in]{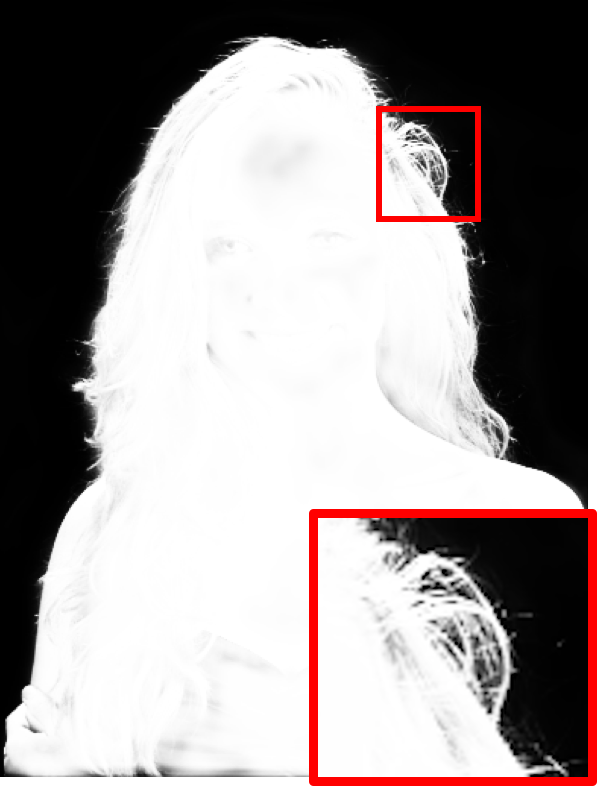}
\includegraphics[width=1in]{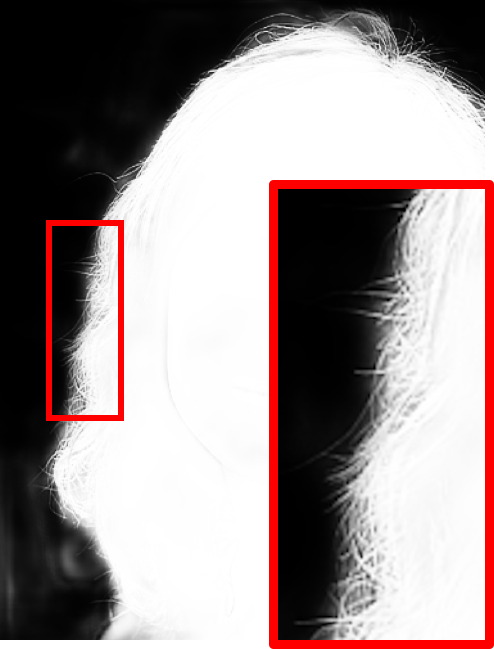}
\end{minipage}
}%
\subfigure[Annotation(KNN)]{
\begin{minipage}[t]{0.33\linewidth}
\centering
\includegraphics[width=1in]{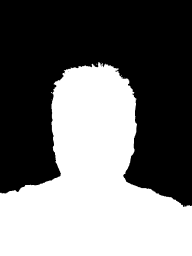}
\includegraphics[width=1in]{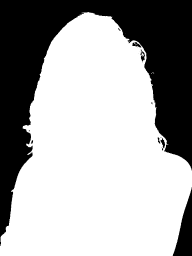}
\includegraphics[width=1in]{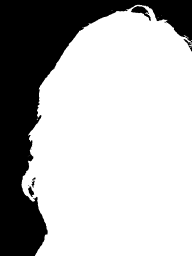}

\end{minipage}%
}%
\centering
\caption{ Details of our outputs}
\label{fig_finer}
\end{figure}

\par In this paper, we propose a specialized portrait segmentation solution by designing  BANet. BANet is an efficient network with only 0.62MB parameters, 
it achieves 43 fps on $512 \times 512$ images with high-quality results which are finer than annotations. Some examples are presented
in Fig\ref{fig_finer} . Our experiments show that BANet generates better results than previous portrait segmentation models and real-time segmentation models
on PFCN+ dataset.


\section{Related Work}
\noindent \textbf{Portrait Segmentation Networks}: PFCN+\cite{shen2016automatic} provides a benchmark for the task of 
portrait segmentation. It calculates firstly an average mask from training dataset, then aligns the average mask 
to each input images to provide prior information. However, the process of alignment requires facial feature points.
So, this method needs an additional landmark detection model, which makes the whole process slow and redundant. BSN\cite{du2017boundary}
proposes a boundary-sensitive kernel to take semantic boundary as a third class and it applies many training tricks such
as multi-scale training and multi-task learning. However, it still lacks fine details and the inference speed is very slow.\\

\noindent \textbf{Image Matting Networks}: Image matting has been applied for portrait image process for a long time. Traditional 
matting algorithms \cite{chuang2001bayesian} \cite{levin2008closed} \cite{chen2013knn} \cite{sun2004poisson} require a user defined trimap which
limits their applications in automatic image process. Some works  \cite{shen2016deep} \cite{chen2018semantic} have proposed to generate trimap by using deep leaning models, 
but they still take trimap generation and detail refinement as two separated stages. In matting tasks, trimap helps to locate regions of interest for alpha-matte.  
Inspired by the function of trimap in matting tasks, we propose a boundary attention
mechanism  to help our BANet focus on boundary areas. Different from previous matting models, we use a two-branch architecture. Our attention map is generated by high-level 
semantic branch, and it is used to guide mining low-level features.    
DIM\cite{xu2017deep} designs a compositional loss which has been proved to be effective in many
matting tasks, but it is time-consuming to prepare large amount of fore-ground, back-ground images and high-quality alpha matte. \cite{Levinshtein2017RealtimeDH} presents a 
gradient-coinstency loss which is able to correct gradient direction on prediction edges, but it is not able to extract richer details. Motivated by that, we propose refine loss to obtain richer detail features on boundary area by using image gradient.\\    

\noindent \textbf{High Speed Segmentation Networks}: ENet \cite{paszke2016enet} is the first semantic segmentation network that achieves real-time performance. It adopts
ResNet \cite{he2016deep} bottleneck structure and reduces channel number in order of acceleration, but ENet loses too much accuracy  as a tradeoff. ICNet\cite{zhao2017icnet} proposes a multi-stream architecture. Three
streams extract features from images of different resolution, and then those features are fused by a cascade feature fusion unit. BiSeNet \cite{yu2018bisenet} uses a two-stream framework to extract context information and spatial information independently. Then it uses a feature fusion module to combine features of two streams. Inspired by their ideas of separate semantic branch and spatial branch, we design a two-stream architecture. Different from previous works, our two branches are not completely separated. In our framework, low-level branch is guided by high-level branch via a boundary attention map.\\

\noindent \textbf{Attention Mechanism}: Attention mechanism makes high-level information to guide low-level feature extraction. SEnet \cite{hu2017squeeze} applies channel 
attention on image recognition tasks and achieves the state-of-the-art. ExFuse \cite{zhang2018exfuse} proposes a Semantic Embedding mechanism to use high-level feature to guide 
low-level feature. DFN \cite{yu2018learning} learns a global feature as attention to revise the process of feature extraction. In PFCN+ \cite{shen2016automatic}, shape channel can be viewed as a kind of spatial-wise attention, aligned mean mask forces model to focus on portrait area.

\begin{figure*}[htbp]
\centering
\includegraphics[scale=0.38]{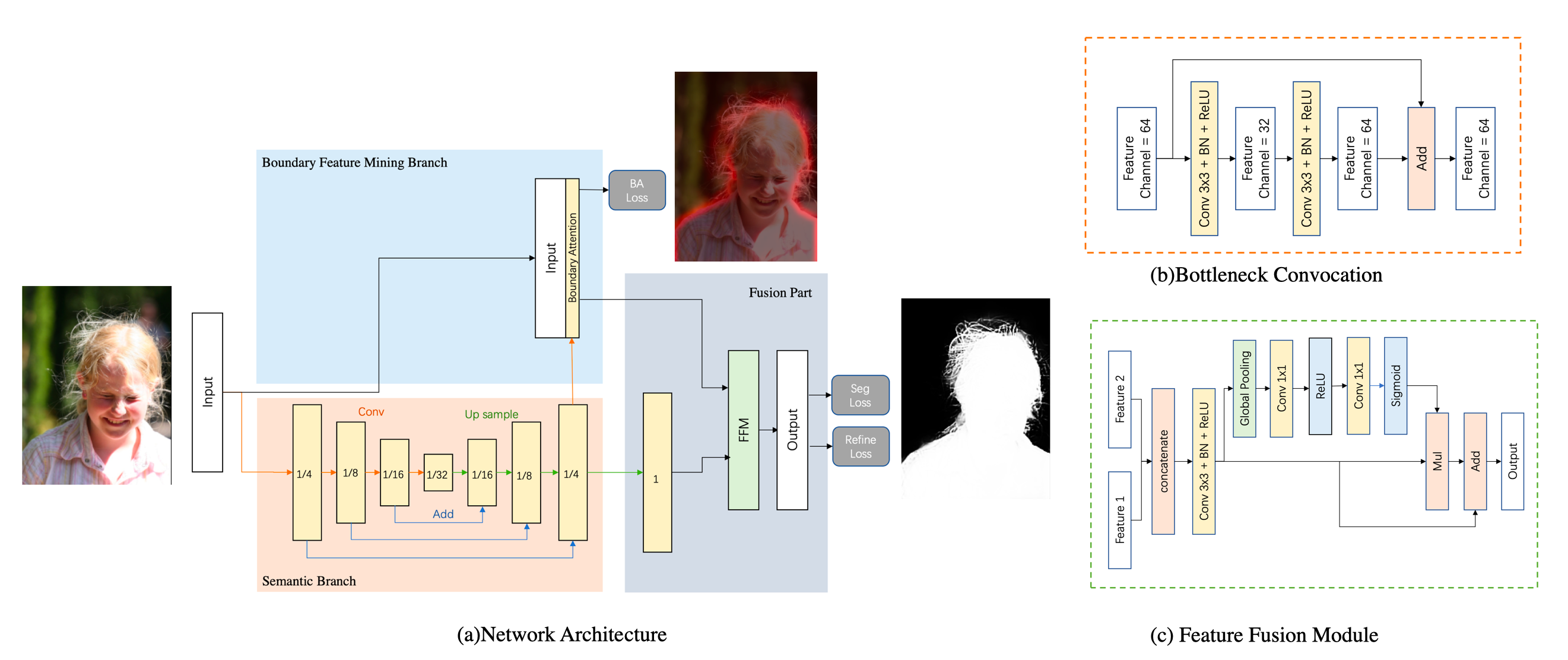} 
\caption{Pipeline of Boundary-Aware Network has three parts. Semantic branch extracts high-level semantic features and produces boundary attention map.
Boundary feature mining branch extracts low-level details. Fusion Part fuses high level features and low-level features by a Feature Fusion Module(c).
In semantic branch, ResNet bottleneck structure(b) is used as basic convolutional block.
}
\label{fig_framework}
\end{figure*}

\section{Our Method}
\par \noindent In the task of portrait segmentation, no-boundary area needs a large receptive field to make prediction with global context information, while 
boundary area needs small receptive field to focus on local feature contrast. Hence these two areas need to be treated independently. In this paper, we propose
a boundary attention mechanism and a weighted loss function to deal with boundary area and no-boundary area separately. 
\par Pipeline of our method is demonstrated as Fig.\ref{fig_framework}. Firstly, color image passes through semantic branch to get 
1/4 size high-level semantic features. Secondly, the output of semantic branch is projected into a one-channel feature map and upsampled to full size
as boundary attention map. This boundary attention map is supervised by boundary attention loss (BA loss). Then, in boundary feature mining branch, we concatenate input 
image with the boundary attention map in order to mining low-level features more goal-directed. Lastly, the fusion part fuses high-level semantic features and 
low-level details to produce fine segmentation results. The final output is supervised by two losses, segmentation loss controls the whole process of 
portrait segmentation and refine loss refines boundary details.

\subsection{ Network Architecture }
\par \noindent \textbf{Semantic Branch}: The objective of semantic
 branch is getting a reliable feature representation. This feature representation is dominated by high-level semantic information. At same time, we hope it can contain some spatial information.  In the task of portrait segmentation, portrait often occupies a
 large  part of image. Hence, large receptive field is required. Semantic branch is a symmetrical fully convolutional structure that follows FCN-4s structure, we utilize multiple downsampling layers to enlarge the receptive field step by step.  As is shown in Fig.\ref{fig_framework}, 
 input image passes through a sequence of convolution layers to get a high-level 1/32 feature map.
 Then, bilinear interpolation is applied to upsample small scale feature maps. We combine features of 1/16,1/8 and 1/4 scale step
 by step by elementwise addition. 
 In order to save computation, we use ResNet bottleneck structure as our basic convolutional block, and we fix the max number of channel
 as 64. The output of semantic branch is a 1/4 size feature map. This feature map has a robust semantic representation because it has large receptive field and it involves 
 4s level spatial information. More detail information is restorated in Boundary Feature Mining Branch.\\

\par
\begin{figure}[htbp]
\centering
\subfigure[Portrait Annotation]{
\begin{minipage}[t]{0.33\linewidth}
\centering
\includegraphics[width=1in]{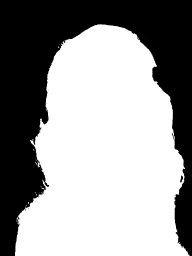}
\includegraphics[width=1in]{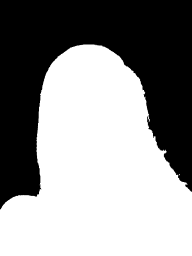}

\end{minipage}%
}%
\subfigure[Boundary Target]{
\begin{minipage}[t]{0.33\linewidth}
\centering
\includegraphics[width=1in]{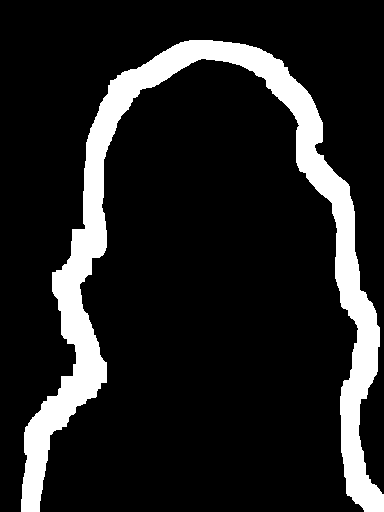}
\includegraphics[width=1in]{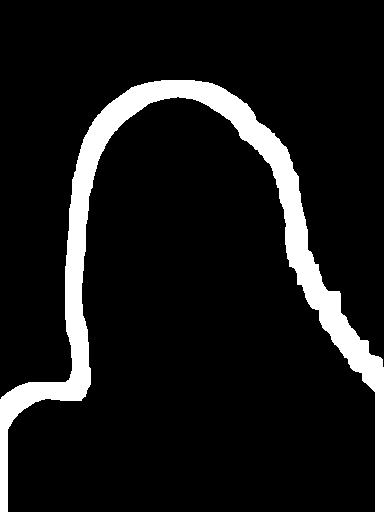}
\end{minipage}%
}%
\subfigure[Boundary Attention]{
\begin{minipage}[t]{0.33\linewidth}
\centering
\includegraphics[width=1in]{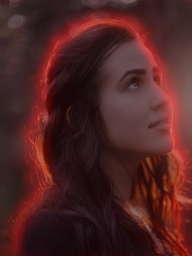}
\includegraphics[width=1in]{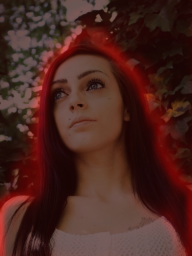}
\end{minipage}
}%
\centering
\caption{ Boundary Attention }
\label{fig_BA}
\end{figure}

 \noindent \textbf{Boundary Feature Mining Branch}: The output of semantic branch is projected to one channel by $1\times1$ conv filter, then it is interpolated to
full size as boundary attention map. BA loss guides this attention map to locate boundary area. Target of boundary area can be generated without manual annotation. As given
in Fig.\ref{fig_BA}.
 Firstly,
we use canny edge detector \cite{canny1986computational} to extract semantic edges on portrait annotation. Considering portrait area varies a lot in different images, it is not 
reasonable to dilate each edge to the same width. So, we dilate the edge of different image with different kernel size as follows:

\begin{equation}
K_{dilation} = \frac{ S_{portrait}}{S_{portrait} + S_{background}}  * W      \label{eq:ratio}
 \end{equation}
 where W is an empirical value representing the canonical width boundary. In our experiment, we set W = 50. 
 \par BA loss is a binary cross-entropy loss, but we don't want our boundary attention map to be binarized because spatial-wise unsmoothness may cause numerical
 unstability.
 Hence, we soften the output of boundary attention map with a Sigmoid function:
\begin{equation}
 S(x) = \frac{1}{1 + e^{-\frac{x}{T}}}       \label{eq:sigmoid}
 \end{equation}
 where T is a temperature which produces a softer probability distribution.
\par Lastly, input image and attention map are concatenated together as a 4-channel image. This 4-channel image passes through a convolution layer
to extract detail information. Different from other multi-stream network architecture that handles each stream independently, our low-level features are guided by high-level features.\\

\par \noindent \textbf{Fusion Part}: Features of semantic branch and boundary feature mining branch are different in level of feature representation. Therefore, we
can't simplely combine them by element-wise addition nor channel-wise concatenation. In Fusion Part, we follow BiSeNet\cite{yu2018bisenet}. Given two features of different
levels, we first concatenate them together then we pass them into a sequence of $3\times3$ convolution, batch normalization and ReLU function. Next, we ulitize global pooling
to calculate a weight vector. This vetor helps feature  re-weight and feature selection. Fig.\ref{fig_framework} (b) shows details of Feature Fusion Module.

\subsection{ Loss Function }
\noindent Our loss function contains three parts as given in Equation \ref{eq:loss}. Where $ L_{seg}$  guides our segmentation 
results,  $ L_{bound}$  supervises boundary attention map, and $ L_{refine}$  refines boundary details.\\

\noindent \textbf{Segmentation Loss}:  Considering the output of portrait segmentation is a one-channel confidence map, binary cross entropy loss is applied as Equation \ref{eq:seg loss}.
where $y_p^s $ represents prediction result of segmentation  and $y_t^s$ stands for segmentation target.
\begin{equation}
 L_{seg} = -[y_{t}^s \cdot log(y_{p}^s ) + (1- y_{t}^s) \cdot log(1-y_{p}^s)  )]   \label{eq:seg loss}
 \end{equation}

\noindent \textbf{Boundary Attention Loss}: Loss function is the same for the boundary attention map, where $y_p^b $ means prediction  of boundary area.
Boundary loss helps our network locate boundary area. At same time, it can be viewed as an intermediate supervision. Extraction of semantic boundary forces
the network to learn a feature with strong inter-class distinction ability.
 
\begin{equation} 
 L_{bound} = -[y_{t}^b \cdot log(y_{p}^b ) + (1- y_{t}^b) \cdot log(1-y_{p}^b )  )] \label{eq:bound loss}
 \end{equation}

\noindent \textbf{Refine Loss} : Refine loss is composed by two parts: $L_{cos}$ and $L_{mag}$. The first part $L_{cos}$ uses cosine distance to 
supervise the gradient direction of segmentation confidence map, $L_{mag}$ brings a constrain on gradient
magnitude to inforce the network produce clear and sharp results.

\par Let $m_{img}$ be the magnitude of gradient of an image, normalized vector $\overrightarrow \nu_{img} = ( g_{img}^x, g_{img}^y)$ denotes
 the direction of gradient of this image. The output of BAnet is a one-channel confidence map of portrait/background dense
 classification. Accordingly, we use $m_{pred}$ and $\overrightarrow \nu_{pred}$ to represent magnitude and direction of gradient for the 
 output confidence map.

\par Inspired by \cite{Levinshtein2017RealtimeDH}, a loss measuring the consistency between image gradient and network output 
is introduced. Our $L_{cos}$ can be formulated as Equation \ref{eq:cos loss}.
\begin{equation} 
\begin{aligned}
L_{cos} &= (1 - | \overrightarrow \nu_{img} \cdot  \overrightarrow \nu_{pred} |) \cdot m_{pred}   \\
& = (1- |g_{img}^x \cdot g_{pred}^x + g_{img}^y \cdot g_{pred}^y |) \cdot m_{pred}     \label{eq:cos loss}
\end{aligned}
\end{equation}

\par While $L_{cos}$ can only deal with cases with simple background. When background is confusing, $L_{cos}$ tends to cause
blurry prediction boundaries. Different from \cite{Levinshtein2017RealtimeDH}, we add a constrain for gradient magnitude as
Equation \ref{eq:mag loss} to force our network to make clear and sharp decision in boundary area. where $\lambda$
is a factor that balances distributional differences between image and output confidence map. In our experiment, $\lambda = 1.5$.
$L_{mag}$ enables BANet to amplify local contrast when the color of foreground is very similar to background.

\begin{equation}{}
 L_{mag} = max( \lambda m_{img} - m_{pred}, 0 )  \label{eq:mag loss}
 \end{equation}

 \par While boundary areas need rich low-level features to obtain detail information, areas far from boundaries
 are mainly guided by high-level semantic features, too much low-level features will disturb the segmentation result. 
 Therefore, we should combine low-level features selectively. Matting models \cite{xu2017deep} \cite{chen2018semantic}
 treat boundary areas and no-boundary areas separately in different stages and then compose two prediction results. Instead of that,
 BANet has an end-to-end architecture, we use a weighted loss function to distinct boundary area and no-boundary area. As shown in Equation \ref{eq: refine loss }  
 refine loss is only applied in boundary area, where $\mathbb M_{bound}$
 is a mask which equals 1 only in boundary area. We set $\gamma_1 = \gamma_2 = 0.5 $ in our experiment.

 \begin{equation}
 L_{refine} = \left(\gamma_1 L_{cos} + \gamma_2 L_{mag}\right) \cdot \mathbb M_{bound} \label{eq: refine loss }
 \end{equation}
\noindent \textbf{Total Loss} : Total loss is formulated as a weighted sum of segmentation loss, boundary attention loss and 
refine loss. Experiments shows $\alpha = 0.6, \beta = 0.3, \gamma = 0.1 $ makes good performance. 
 \begin{equation}
 L = \alpha L_{seg} + \beta L_{bound} + \gamma L_{refine} \label{eq:loss}
 \end{equation}

\subsection{Gradient Calculation}

\begin{figure}[htbp]
\centering
\subfigure[Image]{
\begin{minipage}[t]{0.25\linewidth}
\centering
\includegraphics[width=0.8in]{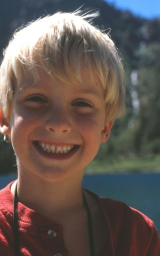}
\end{minipage}%
}%
\subfigure[$M_{img}$]{
\begin{minipage}[t]{0.25\linewidth}
\centering
\includegraphics[width=0.8in]{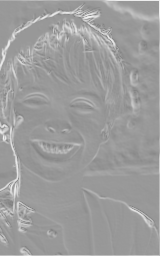}
\end{minipage}%
}%
\subfigure[Prediction]{
\begin{minipage}[t]{0.25\linewidth}
\centering
\includegraphics[width=0.8in]{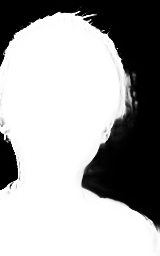}
\end{minipage}
}%
\subfigure[$M_{pred}$]{
\begin{minipage}[t]{0.25\linewidth}
\centering
\includegraphics[width=0.8in]{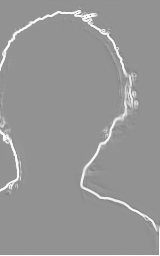}
\end{minipage}
}%
\centering
\caption{ Gradient Calculation }
\label{fig_gadient}
\end{figure}

$L_{refine}$ needs to calculate image gradient, we design a Gradient Calculation Layer(GCL) to compute image gradient on GPU.
Sobel operator \cite{sobel1972camera} is used as convolutional filter in GCL.

\begin{equation}
G_x=
\left[ \begin{array}{ccc}
+1 & 0 & -1\\
+2 & 0 & -2\\
+1 & 0 & -1
\end{array} 
\right ] * I,
G_y=
\left[ \begin{array}{ccc}
+1 & +2 & +1\\
0 & 0 & 0\\
-1 & -2 & -1
\end{array} 
\right ] * I
\end{equation}
where * denotes convolution operation. Gradient magnitude M  and gradient vector $\overrightarrow{\nu}$ are computed as follows:

\begin{equation}
M = \sqrt{G_x^2 + G_y^2 } 
\end{equation}

\begin{equation}
\overrightarrow{\nu} =   \left( \frac{G_x}{M} \frac{G_y}{M} \right)
\end{equation}
Considering original image and prediction result have different distribution, before passed through GCL, input is normalized as:
\begin{equation}
I_{norm} =   \frac{I - I_{min}}{ I_{max} - I_{min}}
\end{equation}
Gradient of original image and output confidence map is presented in Fig.\ref{fig_gadient}.

\section{Experiment Details}
\par \noindent \textbf{Dataset:} PFCN+ \cite{shen2016automatic} dataset contains 1700 KNN annotated portrait images. 1400 images for training
and 300 images for test. Supervise.ly contains more than 4000 polygon annotated human segmentation images, but those images are blended by portrait,
full-body-shot and multi-person photo. We utilize Supervise.ly dataset to pretrain our model in order to get a better initialization, and then we fine-tune 
our model on PFCN+ training set, we use PFCN+ test set for evaluation.\\
\par \noindent \textbf{Data Augmentation:} Online augmentation is implemented during the process of loading data, so that
every training sample is unique. Random rotation in range of [-45\grad ,45\grad  ], random flip, random lightness of range
[0.7,1.3] are used in order to improve our network’s ability of generalization.\\
\par \noindent \textbf{Training Strategy:} Considering that  $L_{refine}$ is only applied on boundary area, it needs a proper boundary attention map to 
help locating boundary. Therefore, we firstly pretrain BANet without $L_{refine}$ on Supervise.ly dataset for 40000 iterations to get a good boundary initialization. Then we fine-tune BANet on PFCN+ dataset with $L_{refine}$ for another 40000 iterations. 
\par
\begin{figure}[htbp]
\centering
\includegraphics[scale=0.5]{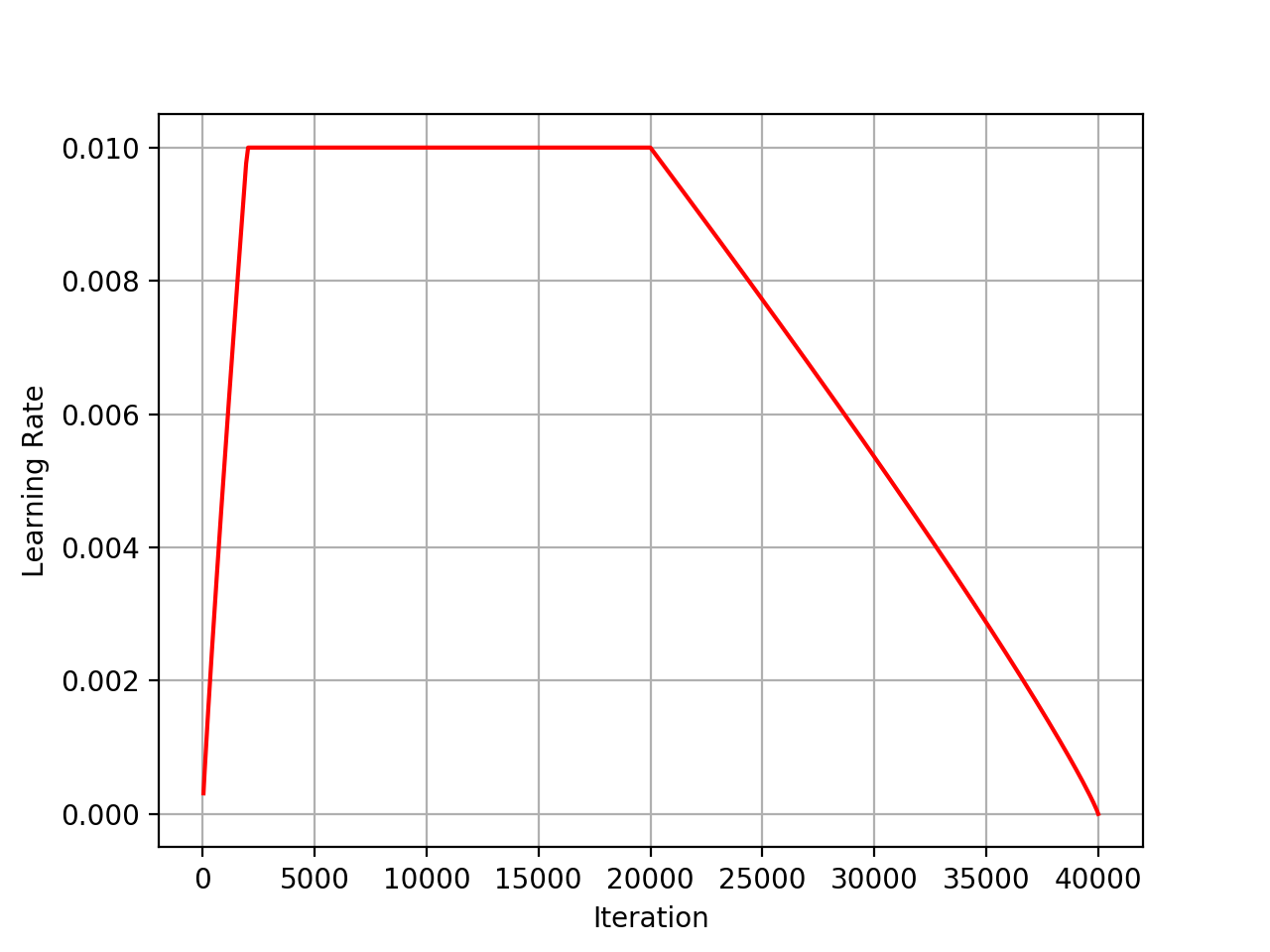} 
\caption{Warming up learning rate
}
\label{fig_warmup}
\end{figure}

 Supervise.ly dataset and PFCN+ dataset have different data distribution. What's more, we add $L_{refine}$ during the process of fine-tuning. Hence, it is not
appropriate to give a big learning rate at the beginning of fine-tuning. We utilize warming-up learning rate which  gives the network a suitable weight initialization before applying big learning rate, thus the network can converge to a better point. Curves of warming-up learning rate can be viewed in Fig.\ref{fig_warmup}
\par Stochastic gradient descent (SGD) optimizer is used to train our model. L2-regularization is applied to all
learnable parameters in order to prevent overfitting. Other hyper parameters for training follows Table.\ref{tabel_hyper}. \\

\begin{table}
\centering
\begin{threeparttable}
\begin{tabular}{|c|c|}
\hline
Hyper Parameter  & Value   \\
\hline\hline
Batch Size  & 16 \\
Weight Decay  & $ 10^{-4}$ \\
Momentum  & 0.9  \\
Max Learning Rate & 0.1 \\
Training Iteration & 4 000 \\
\hline
\end{tabular}
\end{threeparttable}
\caption{ Hyper Parameter
}
\label{tabel_hyper}
\end{table}

\par \noindent \textbf{Comparison Experiments:} Series of comparison experiments are implemented on some real-time 
segmentation networks and portrait segmentation networks. For control variates, the same training strategy and 
hyper parameters are used for all the models. Firstly, these models are pretrained on Supervise.ly dataset, then they
get fine-tuned on PFCN+ dataset. Experiment result is demonstrated in Table.\ref{tabel_result}. Mean IoU is measured as criterion of segmentation accuracy.
Inference speed is measured on a single GTX 1080 Ti with input size $512 \times 512 $. Parameter number evaluates the feasibility of implementing these models
on edge devices such as mobile phone. 
\par In order to prove our BANet performs better than other portrait segmentation networks with the same order of parameter number, we have also implemented a larger version of BANet with 512 channels, called BANet-512. BANet-64 means the version of BANet with 64 channels.\\

\par
\begin{figure*}[htbp]
\centering
\subfigure[Image]{
\begin{minipage}[t]{0.14\linewidth}
\centering
\includegraphics[width=\textwidth]{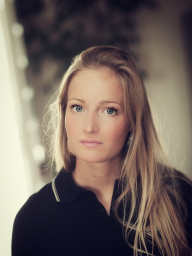}
\includegraphics[width=\textwidth]{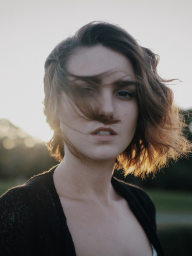}
\includegraphics[width=\textwidth]{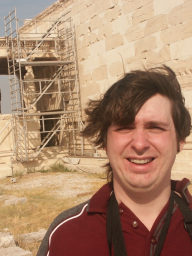}
\includegraphics[width=\textwidth]{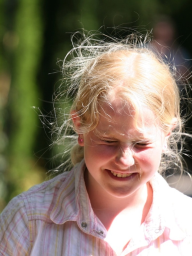}
\includegraphics[width=\textwidth]{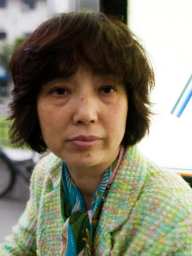}
\includegraphics[width=\textwidth]{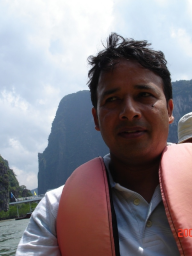}
\end{minipage}%
}%
\subfigure[BiSeNet \cite{yu2018bisenet}]{
\begin{minipage}[t]{0.14\linewidth}
\centering
\includegraphics[width=\textwidth]{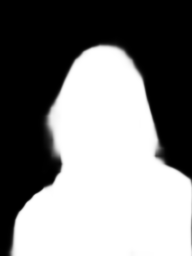}
\includegraphics[width=\textwidth]{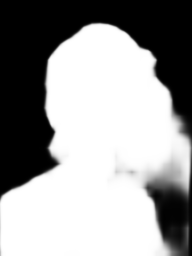}
\includegraphics[width=\textwidth]{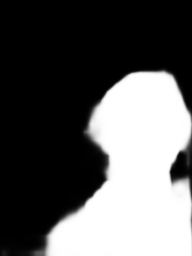}
\includegraphics[width=\textwidth]{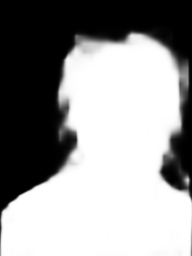}
\includegraphics[width=\textwidth]{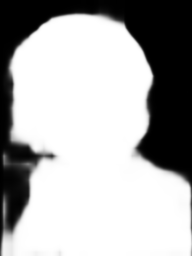}
\includegraphics[width=\textwidth]{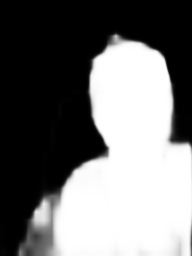}
\end{minipage}%
}%
\subfigure[EDAnet \cite{lo2018efficient}]{
\begin{minipage}[t]{0.14\linewidth}
\centering
\includegraphics[width=\textwidth]{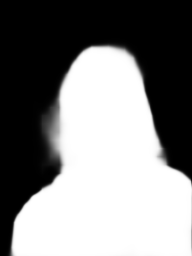}
\includegraphics[width=\textwidth]{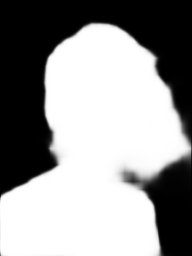}
\includegraphics[width=\textwidth]{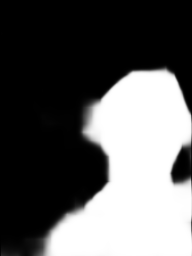}
\includegraphics[width=\textwidth]{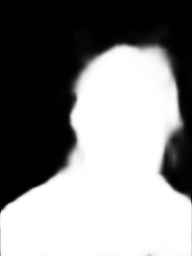}
\includegraphics[width=\textwidth]{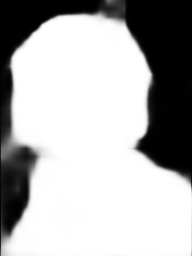}
\includegraphics[width=\textwidth]{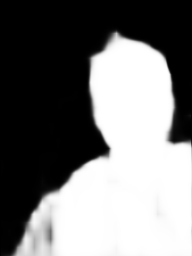}
\end{minipage}%
}%
\subfigure[PFCN+ \cite{shen2016automatic}]{
\begin{minipage}[t]{0.14\linewidth}
\centering
\includegraphics[width=\textwidth]{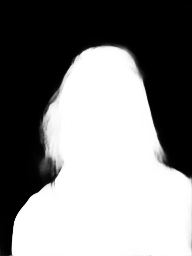}
\includegraphics[width=\textwidth]{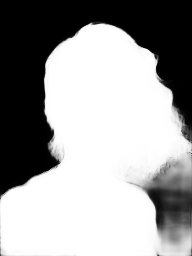}
\includegraphics[width=\textwidth]{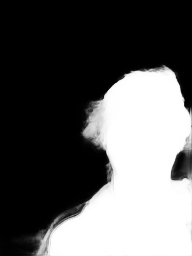}
\includegraphics[width=\textwidth]{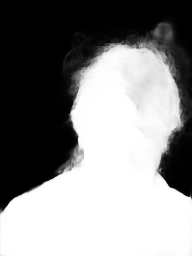}
\includegraphics[width=\textwidth]{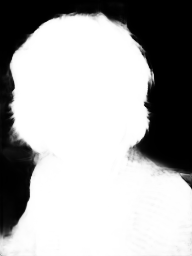}
\includegraphics[width=\textwidth]{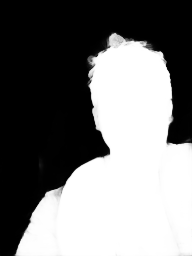}
\end{minipage}%
}%
\subfigure[U-Net \cite{ronneberger2015u}]{
\begin{minipage}[t]{0.14\linewidth}
\centering
\includegraphics[width=\textwidth]{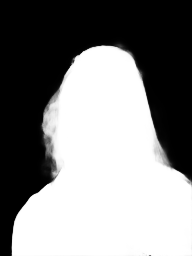}
\includegraphics[width=\textwidth]{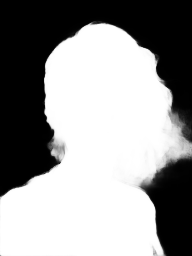}
\includegraphics[width=\textwidth]{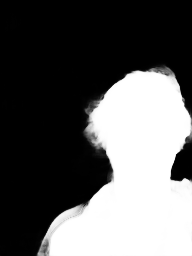}
\includegraphics[width=\textwidth]{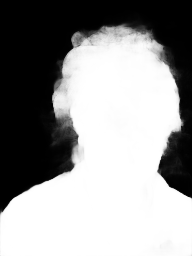}
\includegraphics[width=\textwidth]{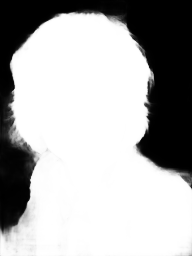}
\includegraphics[width=\textwidth]{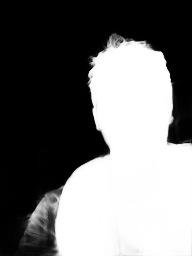}

\end{minipage}%
}%
\subfigure[BANet-64]{
\begin{minipage}[t]{0.14\linewidth}
\centering
\includegraphics[width=\textwidth]{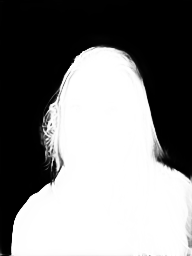}
\includegraphics[width=\textwidth]{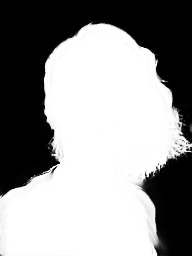}
\includegraphics[width=\textwidth]{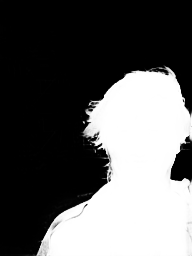}
\includegraphics[width=\textwidth]{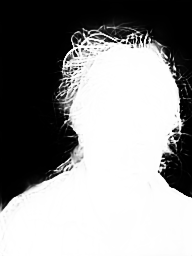}
\includegraphics[width=\textwidth]{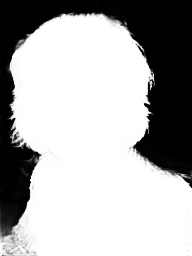}
\includegraphics[width=\textwidth]{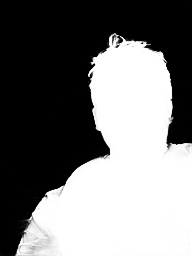}
\end{minipage}%
}%
\subfigure[Annotation(KNN)]{
\begin{minipage}[t]{0.14\linewidth}
\centering
\includegraphics[width=\textwidth]{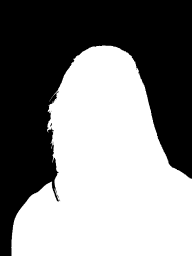}
\includegraphics[width=\textwidth]{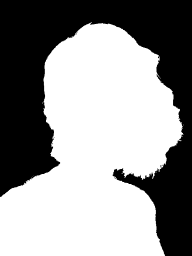}
\includegraphics[width=\textwidth]{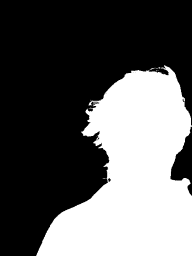}
\includegraphics[width=\textwidth]{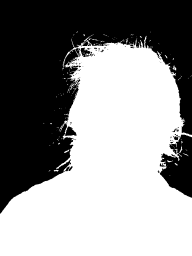}
\includegraphics[width=\textwidth]{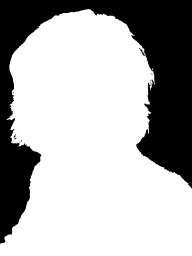}
\includegraphics[width=\textwidth]{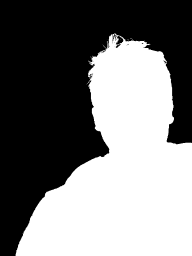}
\end{minipage}%
}%
\centering
\caption{ Comparison results. Traditional Real-time segmentation models like BiSeNet and EDAnet have a huge gap to our work in the task of portrait segmentation.
PFCN+ and U-Net have a higher mean IoU than our result but they have obviously coarser boundaries. Mean IoU of our result is limited by target quality.}
\label{fig_comparison}
\end{figure*}

 \noindent \textbf{Result Analysis:}  As shown in Fig.\ref{fig_comparison}. Confidence maps of BiSeNet and EDAnet have vague boundaries, they lack low-level information because they are upsampled
directly by bilinear interpolation. PFCN+ has a better semantic representation with the help of prior information provided by shape channel. U-Net get a better boundary 
because it combines much low-level information. However, U-Net and PFCN+  are still not able to extract fine details like hairs. With the help of boundary attention map
and refine loss, BANet-64 is able to produce high-quality results with fine details. In some cases, BANet produces segmentation results which are finer than KNN-annotations.
Considering mean IoU measures the similarity between output and annotation, even though PFCN+ and U-Net get a higher mean IoU than BANet-64, we can't conclude BANet-64 is less
accurate than them. While with same order of parameter number, BANet-512 surpasses the other models.
\par What's more,  PFCN+ has more than 130 MB parameters while our BANet-64 has only 0.62 MB parameters, which makes it possible to be implemented on mobile phone. As for inference speed, BANet-64 achieves 43 FPS on $512 \times 512$ images and 115 FPS on $256 \times 256$ images, so it can be applied in real-time image processing.

\begin{table}
\centering
\begin{threeparttable}
\begin{tabular}{|l|c|c|c|}
\hline
model  & mIoU (\%) & speed(fps) & parameter(MB) \\
\hline\hline
ENet\cite{paszke2016enet}  & 88.24 & 46 & 0.35 \\
BiSeNet \cite{yu2018bisenet}  & 93.91 & 59 & 27.01 \\
EDAnet\cite{lo2018efficient}  & 94.90 & 71 & 0.68 \\
U-Net\cite{ronneberger2015u}  & 95.88 & 37 & 13.40 \\
PFCN+\cite{shen2016automatic} & 95.92 & 8 & 134.27 \\
\hline
BANet-64 &95.80& 43 & 0.62 \\
BANet-512 &96.13& 29 & 12.75 \\
\hline
\end{tabular}

\end{threeparttable}
\caption{Comparison results: All experiments are implemented on Pytorch\cite{paszke2017automatic} framework with input images of size
 $512 \times 512$ . Inference speed is measured on a single GTX 1080 Ti.
}
\label{tabel_result}
\end{table}

\begin{figure}[htbp]
\centering
\subfigure[Image]{
\begin{minipage}[t]{0.25\linewidth}
\centering
\includegraphics[width=0.8in]{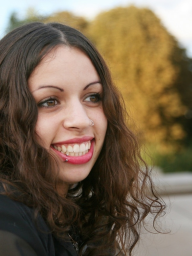}
\includegraphics[width=0.8in]{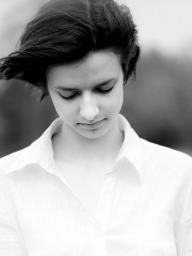}
\includegraphics[width=0.8in]{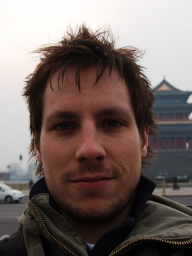}

\end{minipage}%
}%
\subfigure[$Ours^1$]{
\begin{minipage}[t]{0.25\linewidth}
\centering
\includegraphics[width=0.8in]{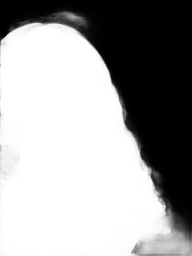}{}
\includegraphics[width=0.8in]{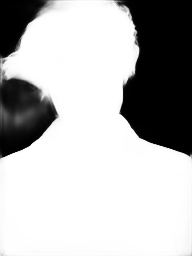}
\includegraphics[width=0.8in] {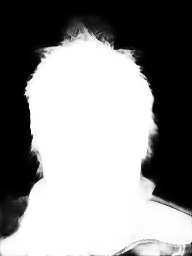}
\end{minipage}%
}%
\subfigure[$Ours^2$]{
\begin{minipage}[t]{0.25\linewidth}
\centering
\includegraphics[width=0.8in]{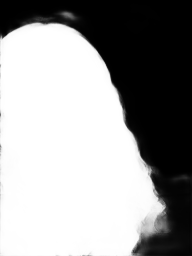}
\includegraphics[width=0.8in]{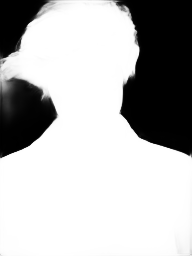}
\includegraphics[width=0.8in]{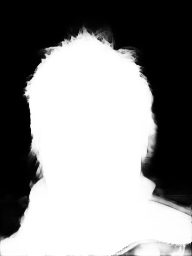}
\end{minipage}
}%
\subfigure[$Ours^3$]{
\begin{minipage}[t]{0.25\linewidth}
\centering
\includegraphics[width=0.8in]{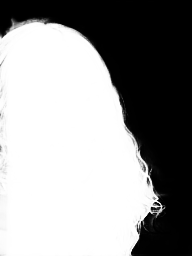}
\includegraphics[width=0.8in]{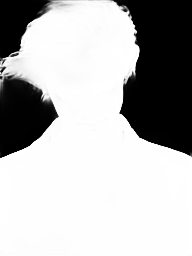}
\includegraphics[width=0.8in]{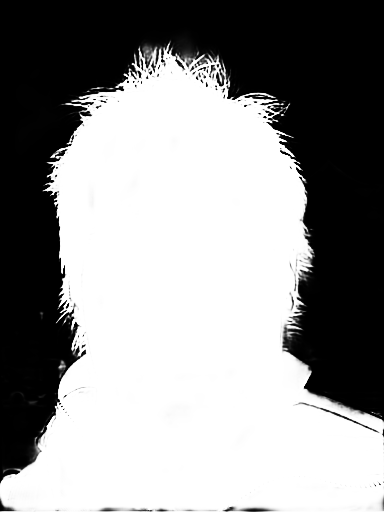}
\end{minipage}
}%
\centering
\caption{ Ablation Study shows boundary attention map brings a better semantic prepresentation and refine loss
brings finer boundaies. }
\label{fig_BANet}
\end{figure}

\begin{table}
\centering
\begin{threeparttable}
\begin{tabular}{|l|c|c|c|}
\hline
model   &Boundary Attention  & Refine Loss & mIoU \\
\hline\hline
Ours\tnote {1} & & & 94.90  \\
Ours\tnote {2} & \checkmark & & 95.20  \\
Ours\tnote {3} & \checkmark  &  \checkmark& 95.80  \\
\hline
\end{tabular}

 \begin{tablenotes}
        \footnotesize
        \item[1] : BANet-64 base model 
   \item[2] : BANet-64 base model  + boundary attention
   \item[3] : BANet-64 base model  + boundary attention + refine loss\\
      \end{tablenotes}
\end{threeparttable}
\caption{ Ablation Study}
\label{ablation}
\end{table}

\section{Ablation Study} 
We have also verified the effectiveness of each part in BANet by implementing three versions. $Ours^1$ represents BANet
base model. In this version, we remove BA loss, refine loss and the little branch of boundary attention map. In boundary feature mining branch, instead of concatenating input image and boundary attention map, we extract low-level features directly on original images.
$Ours^2$ means BANet with boundary attention map.  $Ours^3$ stands for BANet with boundary attention map
and refine loss.
Table \ref{ablation} shows that boundary attention map and refine loss brings 0.3\% and 0.6\% improvements respectively. Fig.\ref{fig_BANet} illustrates
that boundary attention map brings a better semantic representation and refine loss makes striking enhancement on boundary quality. BANet base model involves low-level feature
maps, so it has strong expression ability for details. However, too much low-level feature in no-boundary area may disturb high-level feature representation. Boundary 
attention map helps BANet focus on boundary area and extract low-level feature selectively. Refine loss enables BANet breaks the limitation of annotation quality to get finer 
segmentation result.

\section{Applications} 
\begin{figure}[htbp]
\centering
\subfigure{
\begin{minipage}[t]{0.33\linewidth}
\centering
\includegraphics[width=1in]{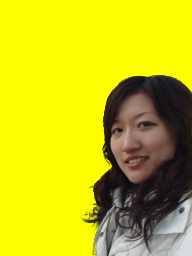}
\includegraphics[width=1in]{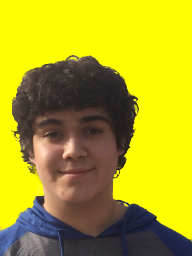}
\includegraphics[width=1in]{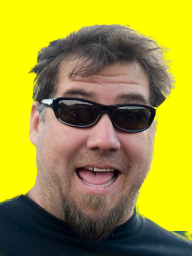}

\end{minipage}%
}%
\subfigure{
\begin{minipage}[t]{0.33\linewidth}
\centering
\includegraphics[width=1in]{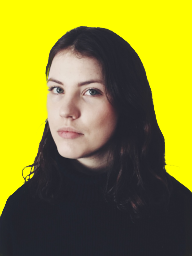}
\includegraphics[width=1in]{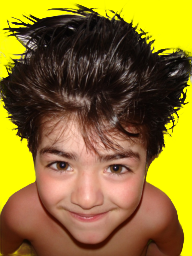}
\includegraphics[width=1in]{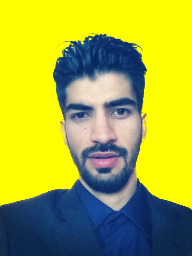}
\end{minipage}%
}%
\subfigure{
\begin{minipage}[t]{0.33\linewidth}
\centering
\includegraphics[width=1in]{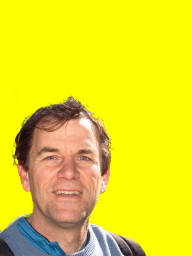}
\includegraphics[width=1in]{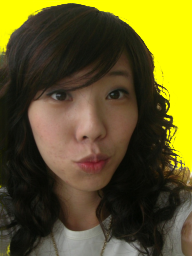}
\includegraphics[width=1in]{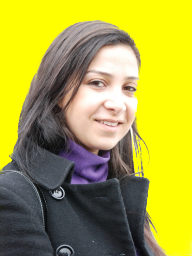}
\end{minipage}
}%
\centering
\caption{ Application on  background changing }
\label{fig_matte}
\end{figure}

Our model can be used in mobile applications for real-time selfie processing. We could directly use the output confidence map of BANet as a soft mask for background changing and image beautification. Compared with other algorithm, our model is able to extract finer features with faster speed and less parameters. Some examples are given in Fig.\ref{fig_matte}.

\section{Conclusion and Limitations} 
In this paper, we propose a Boundary-Aware Network for fast and high-accuracy portrait segmentation. BANet is the first segmentation network that learns to produce results
with high-quality boundaries which are finer than annotations. What's more, light network architecture and high inference speed make it feasible to utilize this model in many 
practical applications. 
\par As for limitations, in order to make BANet lighter and faster, we have removed too much channels in semantic branch. As a tradeoff, our model's ability of feature representation is limited. So,  BANet has difficulty in learning more complicated cases such as multi-person shot and single person portrait with occlusion. 
\par At present, BANet is trained on only 1400 portrait images, it is far from enough if we want to utilize this model on practical applications. In future works, we will annotate more segmentation targets and we will change backgrounds of existing images to get more training data. With the support of a large training dataset,  we will try to
apply BANet on mobile phone.



{\small
\bibliographystyle{ieee}
\bibliography{egbib}

\begin{thebibliography}{10}\itemsep=-1pt

\bibitem{caesar2016coco}
H.~Caesar, J.~Uijlings, and V.~Ferrari.
\newblock Coco-stuff: Thing and stuff classes in context.
\newblock {\em CoRR, abs/1612.03716}, 5:8, 2016.

\bibitem{canny1986computational}
J.~Canny.
\newblock A computational approach to edge detection.
\newblock {\em IEEE Transactions on pattern analysis and machine intelligence},
  (6):679--698, 1986.

\bibitem{chen2018deeplab}
L.-C. Chen, G.~Papandreou, I.~Kokkinos, K.~Murphy, and A.~L. Yuille.
\newblock Deeplab: Semantic image segmentation with deep convolutional nets,
  atrous convolution, and fully connected crfs.
\newblock {\em IEEE transactions on pattern analysis and machine intelligence},
  40(4):834--848, 2018.

\bibitem{chen2018semantic}
Q.~Chen, T.~Ge, Y.~Xu, Z.~Zhang, X.~Yang, and K.~Gai.
\newblock Semantic human matting.
\newblock In {\em 2018 ACM Multimedia Conference on Multimedia Conference},
  pages 618--626. ACM, 2018.

\bibitem{chen2013knn}
Q.~Chen, D.~Li, and C.-K. Tang.
\newblock Knn matting.
\newblock {\em IEEE transactions on pattern analysis and machine intelligence},
  35(9):2175--2188, 2013.

\bibitem{chuang2001bayesian}
Y.-Y. Chuang, B.~Curless, D.~H. Salesin, and R.~Szeliski.
\newblock A bayesian approach to digital matting.
\newblock In {\em null}, page 264. IEEE, 2001.

\bibitem{cordts2016cityscapes}
M.~Cordts, M.~Omran, S.~Ramos, T.~Rehfeld, M.~Enzweiler, R.~Benenson,
  U.~Franke, S.~Roth, and B.~Schiele.
\newblock The cityscapes dataset for semantic urban scene understanding.
\newblock In {\em Proceedings of the IEEE conference on computer vision and
  pattern recognition}, pages 3213--3223, 2016.

\bibitem{du2017boundary}
X.~Du, X.~Wang, D.~Li, J.~Zhu, S.~Tasci, C.~Upright, S.~Walsh, and L.~Davis.
\newblock Boundary-sensitive network for portrait segmentation.
\newblock {\em arXiv preprint arXiv:1712.08675}, 2017.

\bibitem{everingham2010pascal}
M.~Everingham, L.~Van~Gool, C.~K. Williams, J.~Winn, and A.~Zisserman.
\newblock The pascal visual object classes (voc) challenge.
\newblock {\em International journal of computer vision}, 88(2):303--338, 2010.

\bibitem{he2016deep}
K.~He, X.~Zhang, S.~Ren, and J.~Sun.
\newblock Deep residual learning for image recognition.
\newblock In {\em Proceedings of the IEEE conference on computer vision and
  pattern recognition}, pages 770--778, 2016.

\bibitem{hu2017squeeze}
J.~Hu, L.~Shen, and G.~Sun.
\newblock Squeeze-and-excitation networks.
\newblock {\em arXiv preprint arXiv:1709.01507}, 7, 2017.

\bibitem{levin2008closed}
A.~Levin, D.~Lischinski, and Y.~Weiss.
\newblock A closed-form solution to natural image matting.
\newblock {\em IEEE transactions on pattern analysis and machine intelligence},
  30(2):228--242, 2008.

\bibitem{Levinshtein2017RealtimeDH}
A.~Levinshtein, C.~Chang, E.~Phung, I.~Kezele, W.~Guo, and P.~Aarabi.
\newblock Real-time deep hair matting on mobile devices.
\newblock {\em CoRR}, abs/1712.07168, 2017.

\bibitem{lo2018efficient}
S.-Y. Lo, H.-M. Hang, S.-W. Chan, and J.-J. Lin.
\newblock Efficient dense modules of asymmetric convolution for real-time
  semantic segmentation.
\newblock {\em arXiv preprint arXiv:1809.06323}, 2018.

\bibitem{long2015fully}
J.~Long, E.~Shelhamer, and T.~Darrell.
\newblock Fully convolutional networks for semantic segmentation.
\newblock In {\em Proceedings of the IEEE conference on computer vision and
  pattern recognition}, pages 3431--3440, 2015.

\bibitem{paszke2016enet}
A.~Paszke, A.~Chaurasia, S.~Kim, and E.~Culurciello.
\newblock Enet: A deep neural network architecture for real-time semantic
  segmentation.
\newblock {\em arXiv preprint arXiv:1606.02147}, 2016.

\bibitem{paszke2017automatic}
A.~Paszke, S.~Gross, S.~Chintala, G.~Chanan, E.~Yang, Z.~DeVito, Z.~Lin,
  A.~Desmaison, L.~Antiga, and A.~Lerer.
\newblock Automatic differentiation in pytorch.
\newblock 2017.

\bibitem{ronneberger2015u}
O.~Ronneberger, P.~Fischer, and T.~Brox.
\newblock U-net: Convolutional networks for biomedical image segmentation.
\newblock In {\em International Conference on Medical image computing and
  computer-assisted intervention}, pages 234--241. Springer, 2015.

\bibitem{shen2016automatic}
X.~Shen, A.~Hertzmann, J.~Jia, S.~Paris, B.~Price, E.~Shechtman, and I.~Sachs.
\newblock Automatic portrait segmentation for image stylization.
\newblock In {\em Computer Graphics Forum}, volume~35, pages 93--102. Wiley
  Online Library, 2016.

\bibitem{shen2016deep}
X.~Shen, X.~Tao, H.~Gao, C.~Zhou, and J.~Jia.
\newblock Deep automatic portrait matting.
\newblock In {\em European Conference on Computer Vision}, pages 92--107.
  Springer, 2016.

\bibitem{sobel1972camera}
I.~Sobel.
\newblock Camera models and machine perception.
\newblock Technical report, Computer Science Department, Technion, 1972.

\bibitem{sun2004poisson}
J.~Sun, J.~Jia, C.-K. Tang, and H.-Y. Shum.
\newblock Poisson matting.
\newblock In {\em ACM Transactions on Graphics (ToG)}, volume~23, pages
  315--321. ACM, 2004.

\bibitem{xu2017deep}
N.~Xu, B.~L. Price, S.~Cohen, and T.~S. Huang.
\newblock Deep image matting.
\newblock In {\em CVPR}, volume~2, page~4, 2017.

\bibitem{yu2018bisenet}
C.~Yu, J.~Wang, C.~Peng, C.~Gao, G.~Yu, and N.~Sang.
\newblock Bisenet: Bilateral segmentation network for real-time semantic
  segmentation.
\newblock {\em arXiv preprint arXiv:1808.00897}, 2018.

\bibitem{yu2018learning}
C.~Yu, J.~Wang, C.~Peng, C.~Gao, G.~Yu, and N.~Sang.
\newblock Learning a discriminative feature network for semantic segmentation.
\newblock {\em arXiv preprint arXiv:1804.09337}, 2018.

\bibitem{zhang2018exfuse}
Z.~Zhang, X.~Zhang, C.~Peng, D.~Cheng, and J.~Sun.
\newblock Exfuse: Enhancing feature fusion for semantic segmentation.
\newblock {\em arXiv preprint arXiv:1804.03821}, 2018.

\bibitem{zhao2017icnet}
H.~Zhao, X.~Qi, X.~Shen, J.~Shi, and J.~Jia.
\newblock Icnet for real-time semantic segmentation on high-resolution images.
\newblock {\em arXiv preprint arXiv:1704.08545}, 2017.

\bibitem{zhao2017pyramid}
H.~Zhao, J.~Shi, X.~Qi, X.~Wang, and J.~Jia.
\newblock Pyramid scene parsing network.
\newblock In {\em IEEE Conf. on Computer Vision and Pattern Recognition
  (CVPR)}, pages 2881--2890, 2017.

\bibitem{zhu2017fast}
B.~Zhu, Y.~Chen, J.~Wang, S.~Liu, B.~Zhang, and M.~Tang.
\newblock Fast deep matting for portrait animation on mobile phone.
\newblock In {\em Proceedings of the 2017 ACM on Multimedia Conference}, pages
  297--305. ACM, 2017.

\end{thebibliography}
}

\end{document}